\documentclass[
twocolumn,
]{ceurart}

\usepackage{listings}

\newcommand{\FCtoFF}{\Lambda_{FC\rightarrow FF}}
\newcommand{\COCOtoFF}{\Lambda_{COCO\rightarrow FF}}
\begin{document}

\copyrightyear{2022}
\copyrightclause{Copyright for this paper by its authors.
  Use permitted under Creative Commons License Attribution 4.0
  International (CC BY 4.0).}

\conference{Human-in-the-loop data curation workshop at ACM CIKM 2022,
  Oct 17--21, 2022, Atlanta, GA}

\title{From fat droplets to floating forests: cross-domain transfer learning using a PatchGAN-based segmentation model}

\author[1]{Kameswara Bharadwaj Mantha}[%
email=manth145@umn.edu,
]
\cormark[1]
\fnmark[1]
\address[1]{School of Physics \& Astronomy, University of Minnesota, Twin Cities, 116 Church St SE, Minneapolis, MN, 55455}

\author[1]{Ramanakumar Sankar}[%
email=rsankar@umn.edu,
]
\cormark[1]
\fnmark[1]

\author[1]{Yuping Zheng}
\author[1]{Lucy Fortson}
\author[2]{Thomas Pengo}
\address[2]{University of Minnesota Informatics Institute, 2231 6th St SE, Minneapolis, MN, 55455}
\author[3]{Douglas Mashek}
\address[3]{Medical School, University of Minnesota, Twin Cities, 420 Delaware Street SE,  Minneapolis, MN, 55455}
\author[3]{Mark Sanders}
\author[4]{Trace Christensen}
\author[4]{Jeffrey Salisbury}
\address[4]{Mayo Clinic, 200 First Street SW, Rochester, MN, 55905}
\author[5]{Laura Trouille}
\address[5]{Adler Planetarium, 1300 S DuSable Lake Shore Dr., Chicago, IL 60605}
\author[6]{Jarrett E. K. Byrnes}
\author[6]{Isaac Rosenthal}
\address[6]{Department of Biology, University of Massachusetts Boston
100 Morrissey Blvd; Boston, MA, 02125}

\author[7]{Henry Houskeeper}
\author[7]{Kyle Cavanaugh}
\address[7]{Department of Geography, University of California Los Angeles, Los Angeles, CA 90095 }

\cortext[1]{Corresponding author.}
\fntext[1]{KBM and RS contributed equally to the majority of this research. Additional authors contributed to specific aspects including initial models, data sets and project or platform development.}

\begin{abstract}
  Many scientific domains gather sufficient labels to train machine algorithms through human-in-the-loop techniques provided by the Zooniverse.org citizen science platform.  As the range of projects, task types and data rates increase, acceleration of model training is of paramount concern to focus volunteer effort where most needed. The application of Transfer Learning (TL) between Zooniverse projects holds promise as a solution. However, understanding the effectiveness of TL approaches that pretrain on large-scale generic image sets vs. images with similar characteristics possibly from similar tasks 
is an open challenge. We apply a generative segmentation model on two Zooniverse project-based data sets: (1) to identify fat droplets in liver cells (FatChecker; FC) and (2) the identification of kelp beds in satellite images (Floating Forests; FF) through transfer learning from the first project. We compare and contrast its performance with a TL model based on the COCO image set, and subsequently  with baseline 
counterparts.  
We find that both the FC and COCO TL models perform better than the baseline cases when using $>75\%$ of the original training sample size. The COCO-based TL model generally performs better than the FC-based one, likely due to its 
generalized features. Our investigations provide important insights into usage of TL approaches on multi-domain data hosted across different Zooniverse projects, enabling future projects to accelerate task completion.
\end{abstract}


\maketitle

\section{Introduction}
Citizen Science has established itself as a valuable method for distributed data analysis enabling research teams from diverse domains to solve problems involving large quantities of data with complexity levels requiring human pattern recognition capabilities \cite{Trouille2019, Fortson2018}. As the largest citizen science platform, Zooniverse.org has enabled over 2.5 million volunteers to provide over half a billion annotations on hundreds of projects across the sciences and humanities. Many of these projects use the resulting labels to train machine learning algorithms  typically training models from scratch e.g., \cite{Beaumont2014, Zevin2017, Norouzzadeh2017, Wright2017, dominguez-sanchez18, Willi2019, Laraia2019, Ranadive2020, Spiers2020}. To accelerate labeling efficiencies across the platform, the Zooniverse human-machine system should take advantage of transfer learning techniques, especially when volunteer engagement is at a premium. 
When applying transfer learning, a new project would require fewer labels from volunteers to achieve the same performance as training a model from scratch. Volunteer labelers would thus be able to focus on tasks more suited to humans such as anomaly detection e.g., \cite{Walmsley2022}.  

Transfer learning (TL) is an established approach, where the feature space from a pretrained model can be transferred to another framework and fine tuned to perform analogous or different tasks. Feature extraction is typically performed using Deep Convolutional Neural Networks (CNNs) such as \cite{he2016deep,simonyan2014very}. Transfer learning generally uses models trained on data that is either ``out-of-domain'' (i.e., training data characteristics are different from data at hand) or ``in-domain'' (data that are similar or closely relatable to the data at hand). Quantifying the gains provided by these different TL approaches is an active area of research, where studies find several factors to be at play that govern its effectiveness: Accuracy and architecture choice of the pretrained model \cite{Kassani22}, robustness of model to input adversarial noise \cite{Hadi20}, and type of task to which the TL is being applied \cite{thenmozhi2019crop}. Recent works (e.g., \cite{Walmsley2022, Willi2019}) have demonstrated that transfer learning from a model pretrained on in-domain data performs better than transfer learning from out-of-domain data. On the other hand, some studies find that TL models based on out-of-domain data (e.g., ImageNet or COCO datasets) perform on par with or better than the in-domain TL models \cite{majurski2019cell,ma2022transfer}. 

In order to leverage the Zooniverse's large library of image-label pairs across multiple domains, there is thus a clear need to better understand the effectiveness of cross-domain transfer learning.
In particular, we are interested in 
the application of transfer learning specifically to projects that share task similarity across a wide range of domains. For example, image segmentation tasks vary across vastly different disciplines, from cell biology to satellite imagery. Frameworks such as the U-Net \cite{ronneberger2015u}, Recurrent Convolutional Networks such as Mask-RCNNs \cite{he17}, and Generative Adversarial Networks (GANs; e.g., \cite{huo2018adversarial, isola2017image}) have been used to perform such object segmentation across multiple domains and data sets. However, robust learning of such segmentation models from scratch often requires large annotated training samples that may not be available (e.g., medical imaging), which can lead to poor generalizability of the learnt features to newer data, even in related domains. 
While Zooniverse can provide these large annotation sets per project,  this comes at the cost of  volunteer effort which we seek to optimize.

In an effort to increase project completion rates, this study investigates potential machine performance gains through transfer learning across domains by leveraging the shared task similarity between Zooniverse projects. 
We use a PatchGAN-based \cite{isola2017image} segmentation model\footnote{\url{https://github.com/ramanakumars/patchGAN/}}  to investigate the effectiveness of segmenting kelp beds from satellite images. Particularly, we test transfer learning from the COCO dataset (i.e., out-of-domain) and microscopy imaging of lipid droplets in liver cells (pseudo-in-domain) and compare them to their corresponding ``trained from scratch'' counterparts. 

\vspace{-2mm}\section{Methods}
In this section, we detail our PatchGAN architecture \cite{isola2017image}, the training and testing data and its preparation, and the description of the five models analyzed in our work.

\vspace{-4mm}\subsection{PatchGAN Framework}
\begin{figure}[h!]
    \centering
    \includegraphics[width=\columnwidth]{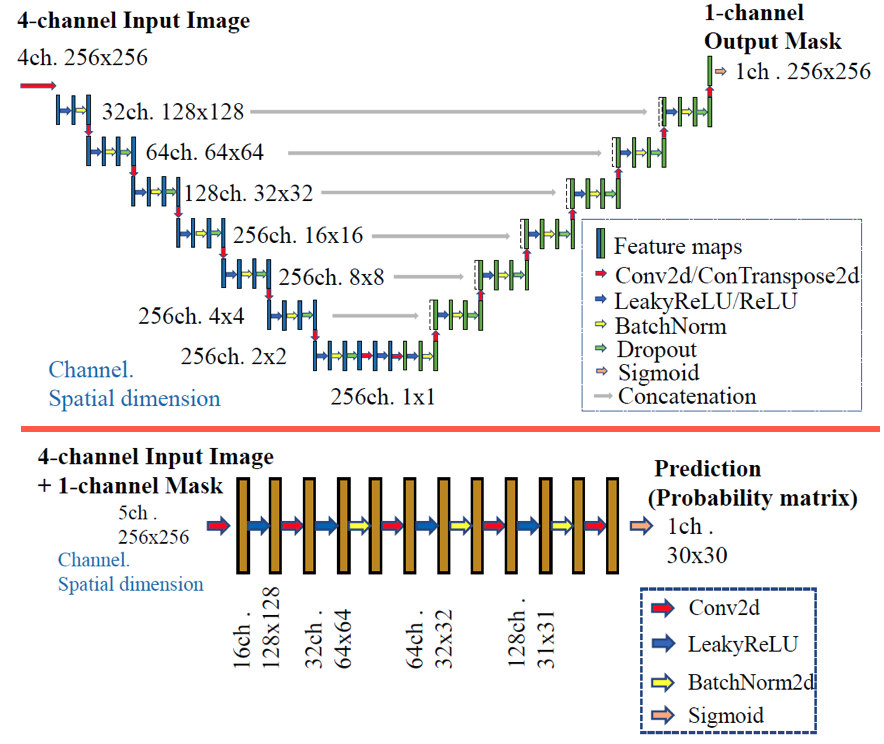}
    \caption{U-Net Generator (top) and Discriminator (bottom) of our PatchGAN framework.}
    \label{fig:generator_architecture}
    \vspace{-3mm}
\end{figure}

The implemented PatchGAN framework is inherited from the Pix2Pix GAN architecture in \cite{isola2017image}, which is a conditional GAN for realizing paired image-to-image translation. 
The PatchGAN architecture consists of a Generator ($G$) and  Discriminator ($D$):

The generator is composed of a U-Net \cite{ronneberger2015u}, a U-shaped encoder-decoder neural network, with skip connections across the bottleneck layer (Figure \ref{fig:generator_architecture}). 
The encoder (decoder) comprises of $6$ downsampling (upsampling) blocks, each consisting of $4\times4$ convolution (transposed convolution), Leaky ReLU activation, and a batch normalization layer. All the blocks in the inner layers of the network also include a dropout layer which omits 50\% of the extracted features during training. The outputs of the transposed convolutions are also concatenated with the corresponding skip connection feature map from the encoder block.

The discriminator is a patch-wise binary classifier that takes a concatenation of the input image and its corresponding ground truth  or generated mask and outputs a $30\times30$ probability matrix. Each unit of this matrix represents a 70 $\times$ 70 patch of the input image, and provides the probability that the patch is real. 

\subsection{Data}\label{sec:data}
For this study, we use three sources for our image-mask pairs: the Floating Forests dataset, Etch-a-Cell dataset and the COCO-stuff. The former two are Zooniverse projects focusing on image segmentation, while the latter represents a generic image dataset that is used in computer vision, representing an out-of-domain dataset compared to the former two. These three data sources represent a diverse feature set on which to perform our transfer learning experiment. Figure\,\ref{fig:gan_results_visualization} shows an example of an image-mask pair from each dataset.

\subsubsection{Floating Forests ($FF$)}
Floating Forests is an ecology-based citizen science project hosted on Zooniverse.org\footnote{\url{https://www.zooniverse.org/projects/zooniverse/floating-forests/}} to identify kelp beds in Landsat imagery. The project presents segments of Landsat data to Zooniverse volunteers, who draw outlines around the kelp beds. These annotations are aggregated using a pixel-by-pixel consensus to create masks of the kelp beds in the corresponding Landsat segments. We use 4 channels from the Landsat data (Blue, Green, Red and near Infrared) to train the patchGAN on the image-mask pairs. This FF data comprises 6,967 ($350\times 350$ pix) image-mask pairs. We pre-process these data such that each pair is cropped into four $256\times 256$ overlapping cutouts, and augment each crop 5 times (rotation and flipping). This resulted in $118,440$ training  and $4180$ testing images.

\subsubsection{Etch-a-Cell: Fat Checker ($FC$)}
Etch-a-Cell: Fat Checker is a cell biology project hosted on Zooniverse.org\footnote{\url{https://www.zooniverse.org/projects/dwright04/etch-a-cell-fat-checker}} to identify lipid droplets in electron microscopy data. The Zooniverse project presents 2D slices of the data to volunteers who annotate the outline of the lipid droplet. The lipid mask is generated by aggregating the annotations by multiple volunteers based on consensus. The data set consists of $2341$ image-mask pairs and each image is $1200\times 1200\,{\rm pix}$ in shape, with 3 channels. We split the sample into $2,106$ training and $235$ testing sets. We transform these images and masks to work with our PatchGAN framework by resizing them to $512 \times 512$ pix and generating five crops (four corners and one center crop). We further augment them by applying three rotations ($90, 180, 270\,{\rm deg}$) per image, yielding augmented training and testing samples of $42120$ and $4700$ images, respectively. 

\subsubsection{COCO-Stuff} The Common Objects in COntext (COCO; \cite{lin2014microsoft}) is a large collection of several real-world images with objects set in various simple to complex scenes, which are annotated by outlines\footnote{\url{https://github.com/nightrome/cocostuff}}. \cite{caesar2018cvpr} further processed the COCO data set to produce dense pixel-wise annotations for them (the COCO-Stuff data set; hereafter COCO). These images and annotated masks vary widely in their shapes, and therefore, we standardize these images by resizing them to a $256\times 256$ pix shape. For our PatchGAN training, we limit the training and testing data to those that host the `person' class. This amounts to $63785$ training and $2673$ testing image-mask pairs. 

\begin{figure}
    \centering
    \includegraphics[width=\columnwidth]{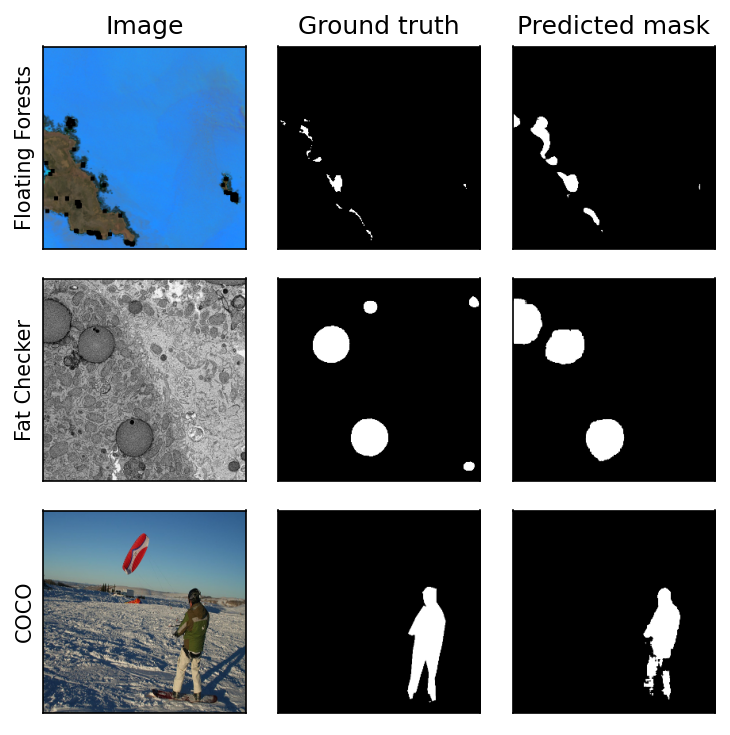}
    \caption{Visualization of example input image, truth mask, and patchGAN predicted output mask.}
    \label{fig:gan_results_visualization}
    \vspace{-5mm}
\end{figure}

\vspace{-2mm}
\subsection{Experimental Design}
In this work, we investigate the potential of cross-domain transfer learning by training $5$ models. The first $3$ models are trained from scratch -- $\Lambda_{FF}$, $\Lambda_{FC}$, and $\Lambda_{COCO}$ -- using $100\%$ of their corresponding data sets $FF$, $FC$, and $COCO$, respectively. Next, we train the $\Lambda_{FC\rightarrow FF}$ and $\Lambda_{COCO\rightarrow FF}$ by transferring the weights from the trained $\Lambda_{FC}$ and $\Lambda_{COCO}$ models to the $\Lambda_{FF}$. By comparing between the baseline $\Lambda_{FF}$ to the transfer learnt models $\Lambda_{FC\rightarrow FF}$ and $\Lambda_{COCO\rightarrow FF}$, we quantify the impact of performing transfer learning on the accelerated learning of the $\Lambda_{FF}$ model from two distinct feature initializations. During this transfer learning exercise, we also vary the amount of training data used from {$10\%$-$100\%$}.\\

\vspace{-2em}
\section{Training \& Results}
In this section, we outline the training strategy and provide details of the hyper parameters. We also present the results of our training and discuss the outcomes of our transfer learning exercise.

\vspace{-1em}
\subsection{Training Strategy}\label{sec:training_strategy}
Our $\Lambda_{FF}$, $\Lambda_{FC}$, and $\Lambda_{COCO}$ models have been trained for $50$ epochs. For the generator, we use the Focal Tversky Loss (FTL; \cite{Abraham2018}), which is a generalized version of the Tversky Loss (TL) defined in terms of the Tversky Index (TI) as:
\begin{equation}
\begin{split}
    TI = \frac{TP}{TP+\alpha FN + \beta FP} \rightarrow TL = (1-TI) \rightarrow FTL = (TL)^{\gamma}, 
\end{split}
\end{equation}
For our training, we use $\alpha = 0.7$ and $\beta = 0.3$. The $\gamma$ parameter controls the non-linearity of the TL with respect to the $TI$, enabling the learning to focus on easier ($\gamma<1$) vs. harder ($\gamma>1$) examples. We use $\gamma=0.75$ during our training. 
For the discriminator optimization, we use the Binary Cross-Entropy ({\it BCE}) loss. Specifically, our total discriminator loss is the average of two components: the discriminator applied on the generated mask (i.e., against a fake label), and applied on the true mask (i.e., the real label).  
For both the generator and discriminator, we use the {\tt Adam} optimizer with an initial learning rate $5\times10^{-4}$ and $1\times10^{-4}$ respectively, decayed exponentially by $\tau = 0.95$, applied every 5 epochs.

\begin{figure*}
    \centering
    \includegraphics[width=1.95\columnwidth]{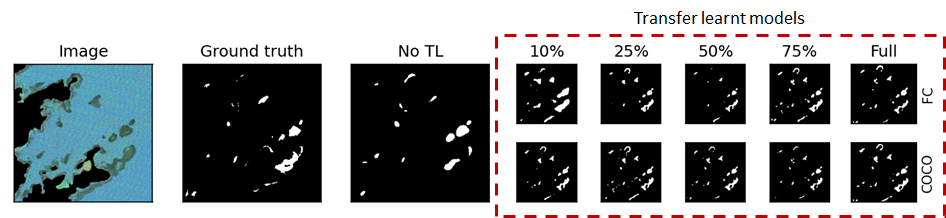}
    \caption{Comparison of generated mask from different model runs on the Floating Forests data, showing different performance gains from transfer learning. }
    \label{fig:FF_output_comparison}
    \vspace{-4mm}
\end{figure*}

\vspace{-2mm}
\subsection{Transfer learning strategy}
For our transfer learning based model training of $\Lambda_{FC\rightarrow FF}$ and $\Lambda_{COCO\rightarrow FF}$, we load the weights of the $\Lambda_{FC}$ and $\Lambda_{COCO}$ models into the freshly initialized $\Lambda_{FF}$ model architecture. To account for the $3$ vs $4$ channel mismatch between the $\Lambda_{COCO}$, $\Lambda_{FC}$ and $\Lambda_{FF}$, we load model layer parameters excluding the input layer.
For each model, we train 5 different versions, using random subsets of $10\%, 25\%, 50\%, 75\%$ and $100\%$ of the full Floating Forests data, to compare TL efficiency gains from having a smaller dataset. For these experiments, we also use only the first 6,967 un-augmented images for re-training. We train the $\Lambda_{FC\rightarrow FF}$ and $\Lambda_{COCO\rightarrow FF}$ models with the same hyper-parameter settings as the aforementioned ``from scratch'' models for $50$ epochs.

\begin{figure}[!h]
    \centering
    \includegraphics[width=\columnwidth]{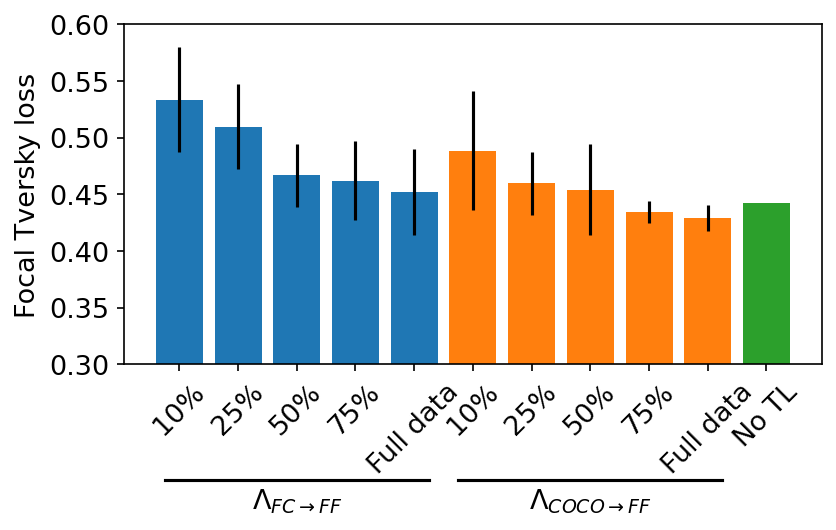}
    \caption{Comparison of mean final loss on Floating Forests validation data across the different models.}
    \label{fig:loss_comparison}
    \vspace{-5mm}
\end{figure}

\subsection{Results and discussion}

We find that our $\Lambda_{FF}$, $\Lambda_{FC}$ and $\Lambda_{COCO}$ generally predict the annotation masks reasonably well (Figure~\ref{fig:gan_results_visualization}), qualitatively matching with the ground truths. 
Figures~\ref{fig:FF_output_comparison} and~\ref{fig:loss_comparison} show our transfer learning results. In Figure~\ref{fig:loss_comparison}, we show our average validation loss for the different model training runs. As expected, larger training samples provide much better performance, but we also find that the model pretrained on the COCO dataset provides noticeably better performance on the Floating Forests data, compared to both $\FCtoFF$ and also $\Lambda_{FF}$. In fact, the $\COCOtoFF$ is able to match the performance of the $\Lambda_{FF}$ model with between 50-75\% of the training Floating Forests dataset.

In Figure~\ref{fig:FF_output_comparison}, we show examples highlighting the difference between the generated masks from $\Lambda_{FF}$ and corresponding masks from $\FCtoFF$ and $\COCOtoFF$. The sharpness of the kelp beds is poorly reconstructed by the $\Lambda_{FF}$ model but is well captured by the transfer learnt models (particularly when training $\Lambda_{COCO\rightarrow FF}$ with more than 75\% of the original data). The transfer learnt models are also better at capturing kelp beds not identified in the original consensus data. For example, both the ground truth and $\Lambda_{FF}$ fail to reveal the kelp beds in the top left of the image, but these are picked up well by the transfer learnt models.

This is likely due to the large diversity of the features in the COCO dataset, making it a much more robust feature extraction network to transfer learn from. Indeed, compared to $\FCtoFF$, the $\COCOtoFF$ model-detected kelp beds are qualitatively better visually (e.g., Figure~\ref{fig:FF_output_comparison}), especially at lower training data sizes. This is likely compounded with the lower feature diversity in both the Floating Forest and Fat Checker data sets, given the fewer number of samples in the training data and low variety in target classes.

\vspace{-1mm}
\subsubsection{Transfer learning approaches for citizen science datasets}
For the Zooniverse platform, this study provides an avenue to build quick access for projects to use machine learning frameworks for simple tasks (e.g., image segmentation), by transfer learning from existing models on a small sample of volunteer annotated data sets. However, despite the results presented here, there are still several key questions which need to be answered: 

{\bf Domain dependency:}
It is unclear how much of the performance gained from COCO was a `global truth'. That is, whether COCO (or similarly diverse datasets) are immediately applicable to out-of-domain data, for all domains, or if there are domain-specific restrictions which allow these performance gains to occur on data such as Floating Forests. This requires more experiments with increasingly different data sets on Zooniverse to investigate the range of performance gains possible. 

{\bf Task dependency: }
Previous studies on transfer learning across domains show significant variations in performance across different task types. For example, image classification tasks (e.g., \cite{Walmsley2022, thenmozhi2019crop}) show lower gains than image segmentation based tasks (e.g., \cite{majurski2019cell}). We need to further investigate the inherent difficulty associated with different tasks on Zooniverse projects, and how effectively they can be transferred between domains. \cite{Walmsley2022}, for example, show that significant boosts to performance is only provided by using in-domain transfer learning. 

{\bf Target data purity: }
For Zooniverse projects, data labels are generally provided by volunteers and are aggregated based on volunteer consensus. In this study, we found that transfer learning can help mitigate data purity effects, since transfer learnt feature extraction models are generally robust to mislabeled data. The extent to which transfer learning models are sensitive to data purity effects needs to be further investigated. 

In conclusion, we find that transfer learning can provide a significant boost to projects that contain similar tasks on Zooniverse. However, the extent to which this can be generalized across the full Zooniverse ecosystem is a question of ongoing study.

\section*{Acknowledgements}
The authors would like to thank the Zooniverse volunteers without whom this work would not have been possible. RS, KM, LF, YZ, LT would like to acknowledge partial support from the National Science Foundation \,  under grant numbers IIS $2006894$ and OAC $1835530$. Partial support by RS, KM, LF, TP, MS, TC, JS is acknowledged through Minnesota Partnership MNP IF\#119.09.

\bibliography{refs}

\end{document}